\documentclass[10pt,twocolumn,letterpaper]{article}
\usepackage{cvpr}
\usepackage{times}
\usepackage{epsfig}
\usepackage{graphicx}
\usepackage{amsmath}
\usepackage{amssymb}
\usepackage{bbm}
\usepackage{upgreek}
\usepackage{algorithm}
\usepackage{algorithmic}

\usepackage{subcaption}

\usepackage[pagebackref=true,breaklinks=true,letterpaper=true,colorlinks,bookmarks=false]{hyperref}

 \cvprfinalcopy 

\def\cvprPaperID{2256} 

\ifcvprfinal\fi
\begin{document}

\title{Visualizing the Decision-making Process in Deep Neural Decision Forest}

\author{Shichao Li and Kwang-Ting Cheng\\
Hong Kong University of Science and Technology \\
Kowloon, Hong Kong\\
{\tt\small nicholas.li@connect.ust.hk, timcheng@ust.hk}
}

\maketitle

\begin{abstract}
   Deep neural decision forest (NDF) achieved remarkable performance on various vision tasks via combining decision tree and deep representation learning. In this work, we first trace the decision-making process of this model and visualize saliency maps to understand which portion of the input influence it more for both classification and regression problems. We then apply NDF on a multi-task coordinate regression problem and demonstrate the distribution of routing probabilities, which is vital for interpreting NDF yet not shown for regression problems. The pre-trained model and code for visualization will be available at \url{https://github.com/Nicholasli1995/VisualizingNDF}
\end{abstract}

\section{Introduction}

Traditional decision trees \cite{7130626, 6909637} are interpretable since they conduct inference by making decisions. An input is routed by a series of splitting nodes and the conclusion is drawn at one leaf node. Training these models follow a local greedy heuristic \cite{7130626, 6909637}, where a purity metric such as entropy is adopted to select the best splitting function from a candidate set at each splitting node. Hand-crafted features were usually used and the model's representation learning ability is limited.

Deep neural decision forest (NDF) \cite{7410529} and its later regression version \cite{123456} formulated a probabilistic routing framework for decision trees. As a result the loss function is differentiable with respect to the parameters used in the splitting functions, enabling gradient-based optimization in a global way. Despite the success of NDF, there is few effort devoted to visualize the decision making process of it. In addition, the deep representation learning ability brought by the soft-routing framework comes with the price of visiting every leaf node in the tree. The model will be more similar to traditional decision tree and more interpretable if few leaf nodes contribute to the final prediction. Fortunately, the desired property was demonstrated by the distribution of routing probabilities in \cite{7410529} for a image classification problem. To our best knowledge, this property has not yet been validated for any regression problem.  

\begin{figure}[t]
	\begin{center}
		\includegraphics[width=1\linewidth]{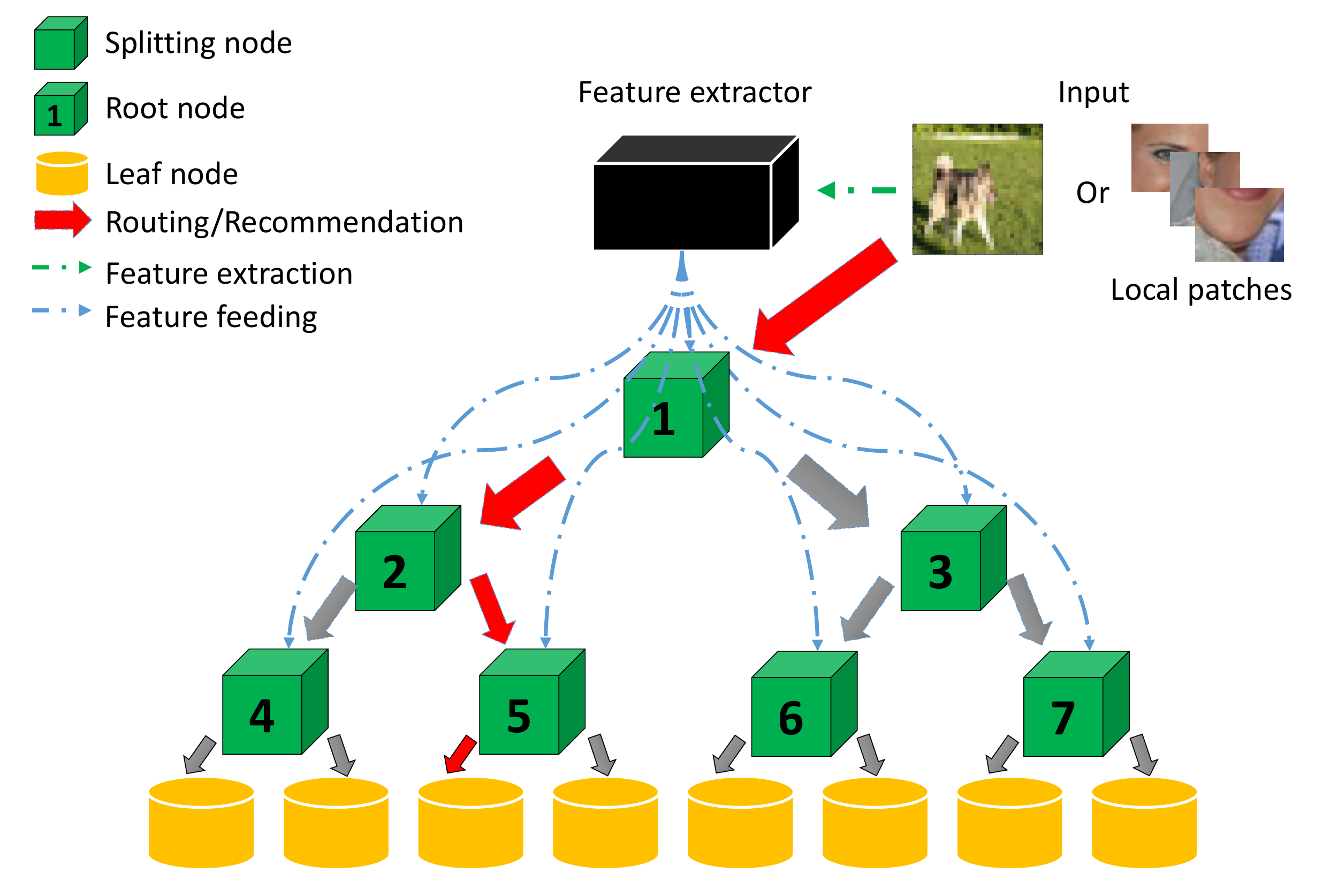}
		\vspace{-0.4in}
	\end{center}
	\caption{Illustration of the decision-making process in deep neural decision forest. Input images are routed (red arrows) by splitting nodes and arrive at the prediction given at leaf nodes. The feature extractor computes deep representation from the input and send it (blue arrows) to each splitting node for decision making. Best viewed in color.}
	\label{model_illus}
\end{figure}

In this paper, we trace the routing of input images and apply gradient-based technique to visualize the important portions of the input that affect NDF's decision-making process. We also apply NDF to a new multi-task regression problem and visualize the distribution of routing probabilities to fill the knowledge blank. In summary, our contributions are:

\begin{itemize}
	\item[1.] We trace the decision-making process of NDF and compute saliency maps to visualize which portion of the input influences it more.
	\item[2.] We utilize NDF on a new regression problem and visualize the distribution of routing probabilities to  validate its interpretability. 
\end{itemize}


\section{Related works}

Traditional classification and regression trees make predictions by decision making, where hand-crafted features \cite{7130626, 6909637} were computed to split the feature space and route the input. Deep neural decision forest (NDF) and its regression variant \cite{123456} were proposed to equip traditional decision trees with deep feature learning ability. Gradient-based method \cite{DBLP:journals/corr/SimonyanVZ13} was adopted to understand the prediction made by traditional deep convolutional neural network (CNN). However, this visualization technique has not yet been applied to NDF. Another orthogonal line of research attempts to learn more interpretable representation \cite{8579018} and organize the inference process into a decision tree \cite{DBLP:journals/corr/abs-1802-00121}. Our work is different from them since it is more of a visualization-based model diagnosis and no other loss function is used in the training phase to drive semantically meaningful feature learning as in \cite{8579018}.



\section{Methodology}
A deep neural decision forest (NDF) is an ensemble of deep neural decision trees. Each tree consists of splitting nodes and leaf nodes. In general each tree can have unconstrained topology but here we specify every tree as full binary tree for simplicity. We index the nodes sequentially with integer $i$ as shown in Figure~\ref{model_illus}. 

A splitting node $\mathcal{S}_{i}$ is associated with a recommendation (splitting) function $\mathcal{R}_{i}$ that extracts deep features from the input $\mathbf{x}$ and gives the recommendation score (routing probability) $s_{i} = \mathcal{R}_{i}(\mathbf{x})$ that the input is recommended (routed) to its left sub-tree.

We denote the unique path from the root node to a leaf node $\mathcal{L}_{i}$ a computation path $\mathcal{P}_{i}$. Each leaf node stores one function $\mathcal{M}_{i}$ that maps the input into a prediction vector $\mathbf{p}_{i} =  \mathcal{M}_{i}(\mathbf{x})$. To get the final prediction $ \mathbf{P}$, each leaf node contributes its prediction vector weighted by the probability of taking its computation path as  
\begin{equation}
\mathbf{P} = \sum_{i\in\mathcal{N}_{l}}w_{i}\mathbf{p}_{i}
\end{equation}
and $\mathcal{N}_{l}$ is the set of all leaf nodes.
The weight can be obtained by multiplying all the recommendation scores given by the splitting nodes along the path. Assume the path $\mathcal{P}_{i}$ consists of a sequence of $q$ splitting nodes and one leaf node as $\{\mathcal{S}_{i_1}^{j_1}, \mathcal{S}_{i_2}^{j_2}, \ldots, \mathcal{S}_{i_q}^{j_q}, {\mathcal{L}_{i}}\}$, where the superscript for a splitting node denotes to which child node to route the input. Here $j_m = 0$ means the input is routed to the left child and $j_m = 1$ otherwise. Then the weight can be expressed as
\begin{equation}
w_{i} = \prod_{m=1}^{q}(s_{i_m})^{\mathbbm{1}(j_m = 0)}(1 - s_{i_m})^{\mathbbm{1}(j_m = 1)}
\end{equation}

Note that the weights of all leaf nodes sum to 1 and the final prediction is hence a convex combination of all the prediction vectors of the leaf nodes. In addition, we assume the recommendation and mapping functions mentioned above are differentiable and parametrized by $\boldsymbol{\uptheta}_{i}$ at node $i$. Then the final prediction is a differentiable function with respect to all the parameters which we omit above to ensure clarity. A loss function defined upon the final prediction can hence be minimized with back-propagation algorithm.

Note here all computation paths will contribute to the final prediction of this model, unlike traditional decision tree where only one path is taken for each input. We believe the model is more interpretable and similar to tradition decision trees when only a few computation paths contribute to the final prediction. This has been shown to be the case for classification problem in \cite{7410529}. Here we also demonstrate the distribution of routing probabilities for a regression problem.

To understand how the input can influence the decision-making of this model, we take the gradient of the routing probability with respect to the input and name it {\itshape decision saliency map (DSM)},

\begin{equation}
DSM = \frac{\partial s_{i}}{\partial \mathbf{x}}
\end{equation}

For classification problem, the prediction vector $\mathbf{p}_{i}$ for each leaf node $\mathcal{L}_{i}$ is a discrete probability distribution vector whose length equals the number of classes. The $y$th entry $\mathbf{p}_{i}(y)$ gives the probability $\mathbb{P}(y|\mathbf{x}) $ that the input $\mathbf{x}$ belongs to class $y$. For regression problems, $\mathbf{p}_{i}$ is also a real-valued vector but the entries do not necessarily sum to 1. The optimization target for classification problems is to minimize the negative log-likelihood loss over the whole training set containing $N$ instances $\mathbb{D} = \{\mathbf{x}_i,y_i\}_{i=1}^{N}$,
$L(\mathbb{D}) = -\sum_{i=1}^{N}\log(\mathbb{P}(y_i|\mathbf{x}_i))$.
For a multi-task regression problem with $N$ instances $\mathbb{D} = \{\mathbf{x}_i,\mathbf{y}_i\}_{i=1}^{N} $, we directly use the squared loss function, 
$L(\mathbb{D}) =\frac{1}{2} \sum_{i=1}^{N}\lvert\lvert\mathbf{P}_{i}-\mathbf{y}_{i}\rvert\rvert^2$.

In the experiment, we use deep CNN to extract features from the input and use sigmoid function to compute the recommendation scores from the features. The network parameters and leaf node prediction vectors are optimized alternately by back propagation and update rule, respectively. Details about the network architectures, training algorithm and hyper-parameter settings can be found in our supplementary materials (included in the GitHub repository). 

\begin{figure*}[t]
	\begin{center}
		\includegraphics[width=1\linewidth]{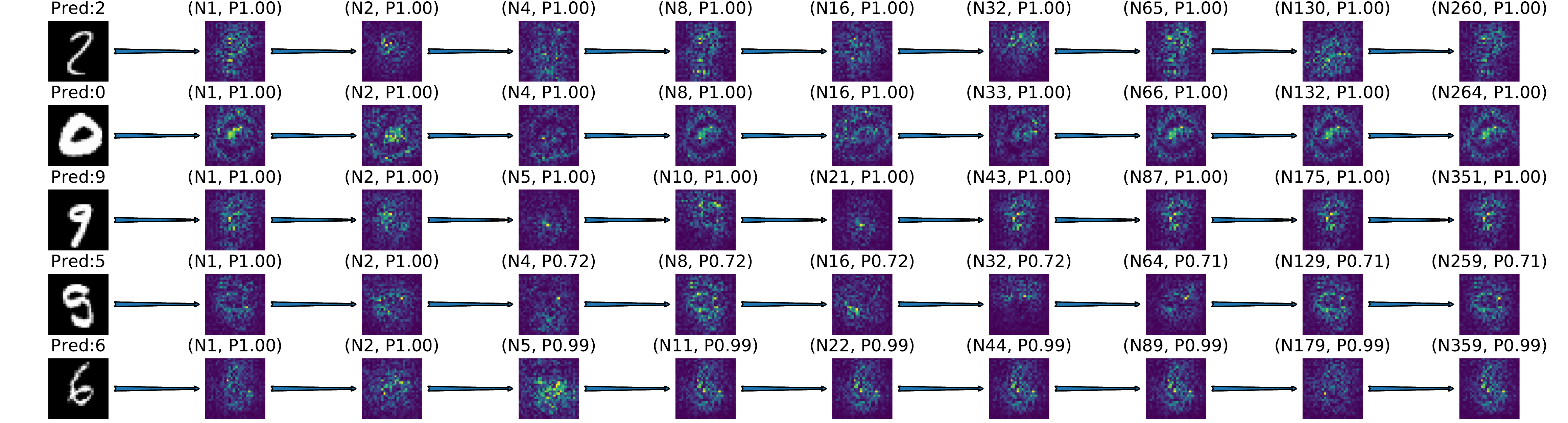}
	\end{center}
	\vspace{-0.2in}
	\caption{Decision saliency maps for MNIST test set. Each row gives the decision-making process of one image, where the left-most image is the input and the others are DSMs along the computation path of the input. Each DSM is computed by taking derivative of the routing probability with respect to the input image. Model prediction is given above the input image and (Na, Pb) means the input arrives at splitting node a with probability b during the decision-making process.}
	
	\label{mnist}
\end{figure*}

\begin{figure*}[t]
	\begin{center}
		\includegraphics[width=1\linewidth]{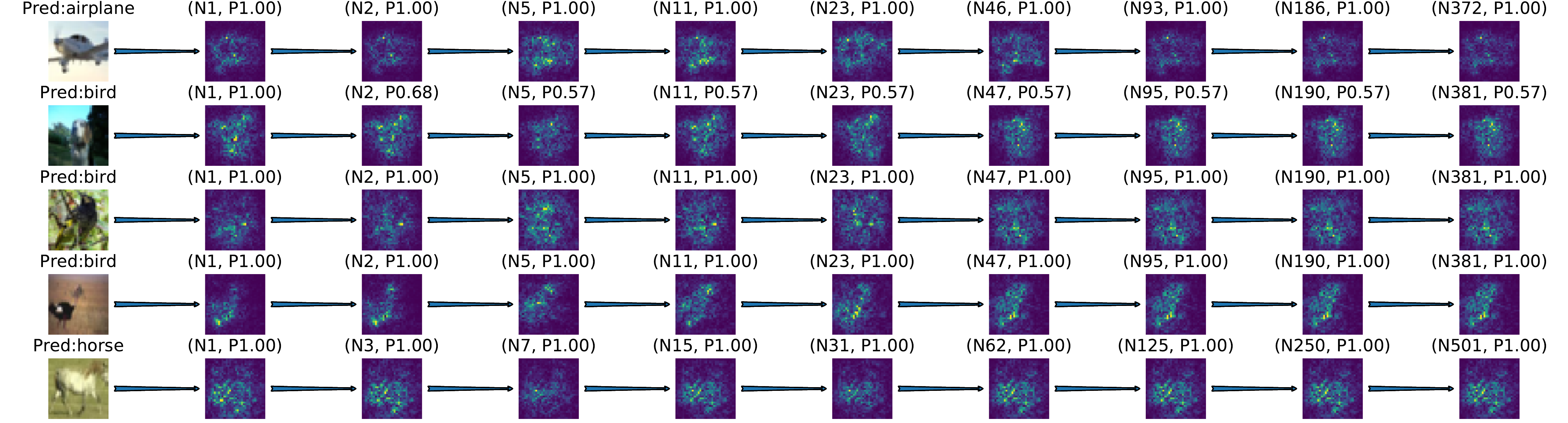}
		\vspace{-0.4in}
	\end{center}
	\caption{Decision saliency maps for CIFAR-10 test set using the same annotation style as Figure~\ref{mnist}. }
	\label{cifar10}
\end{figure*}

\section{Experiments}
\subsection{Classification for MNIST and CIFAR-10}
Standard datasets provided by PyTorch\footnote{\url{https://pytorch.org/docs/0.4.0/_modules/torchvision/datasets}} are used. We use one full binary tree of depth 9 for both datasets, but the complexity of the feature extractor for CIFAR-10 is higher. Adam optimizer is used with learning rate specified as 0.001. Test accuracies for different datasets and feature extractors are shown in Table~\ref{accuracy}. We record the computation path for each test image that has the largest probability been taken, and compute DSMs for some random samples as shown in Fig.~\ref{mnist} and Fig.~\ref{cifar10}. The tree is very decisive as indicated by the probability of arriving at each splitting node. In addition, the foreground usually affect the decision more as expected and also similar to \cite{DBLP:journals/corr/SimonyanVZ13}. Interestingly, the highlight (yellow dots) for different DSMs along the computation path vary a lot for some examples. This means the network is trying to look at different regions of the input while deciding how to route the input. Another interesting observation is that the model mis-classify dog as bird when it is not certain about its decision.

\begin{figure}
	\begin{center}
		\includegraphics[width=1\linewidth]{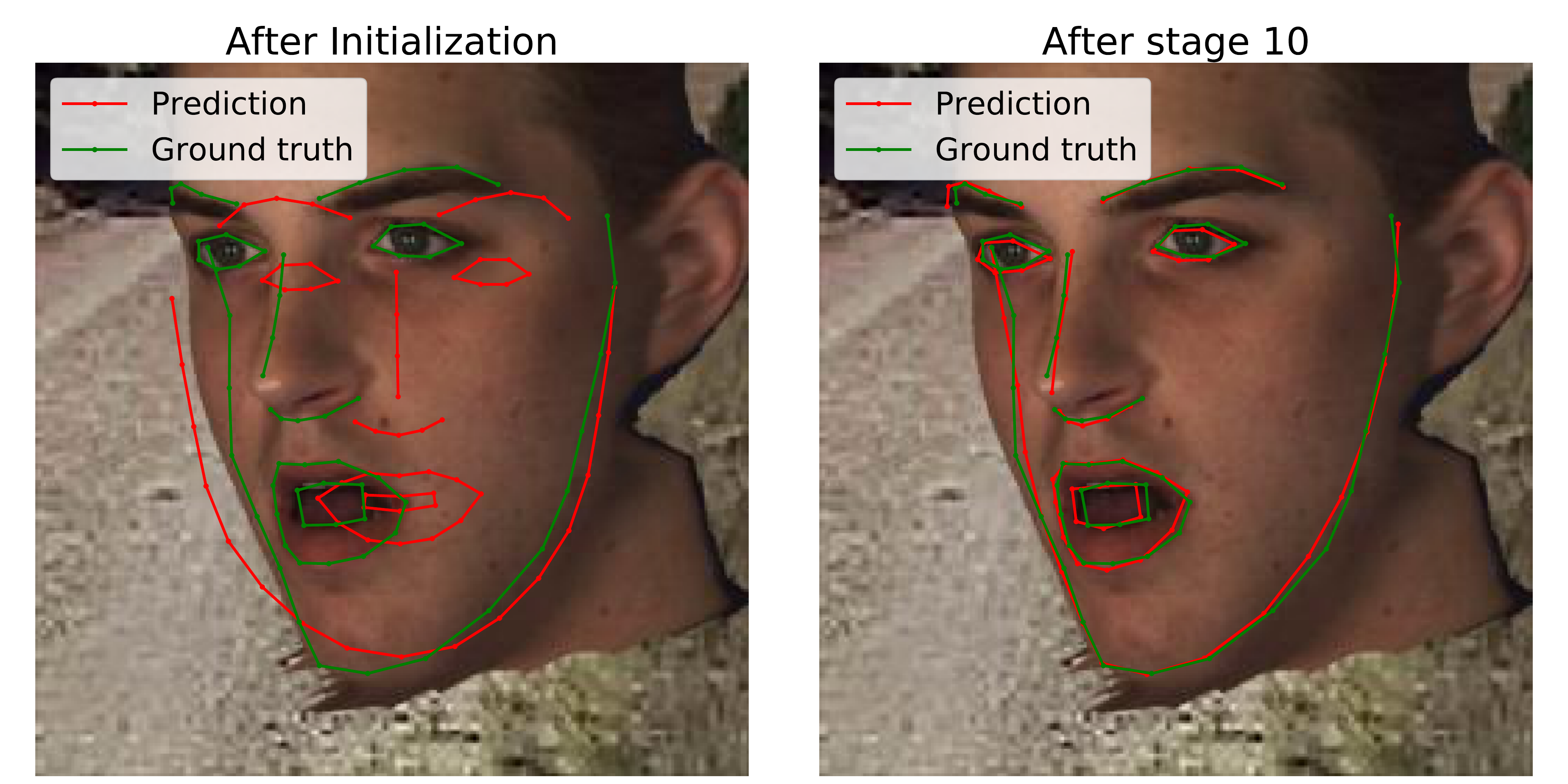}
	\end{center}
	\vspace{-0.2in}
	\caption{Face alignment using a cascade of NDFs. A coarse shape initialization can be updated to well fit the ground truth after 10 stages. Best viewed in color.}
	\label{result}
\end{figure}

\begin{figure}
	\begin{center}
		\includegraphics[width=1\linewidth]{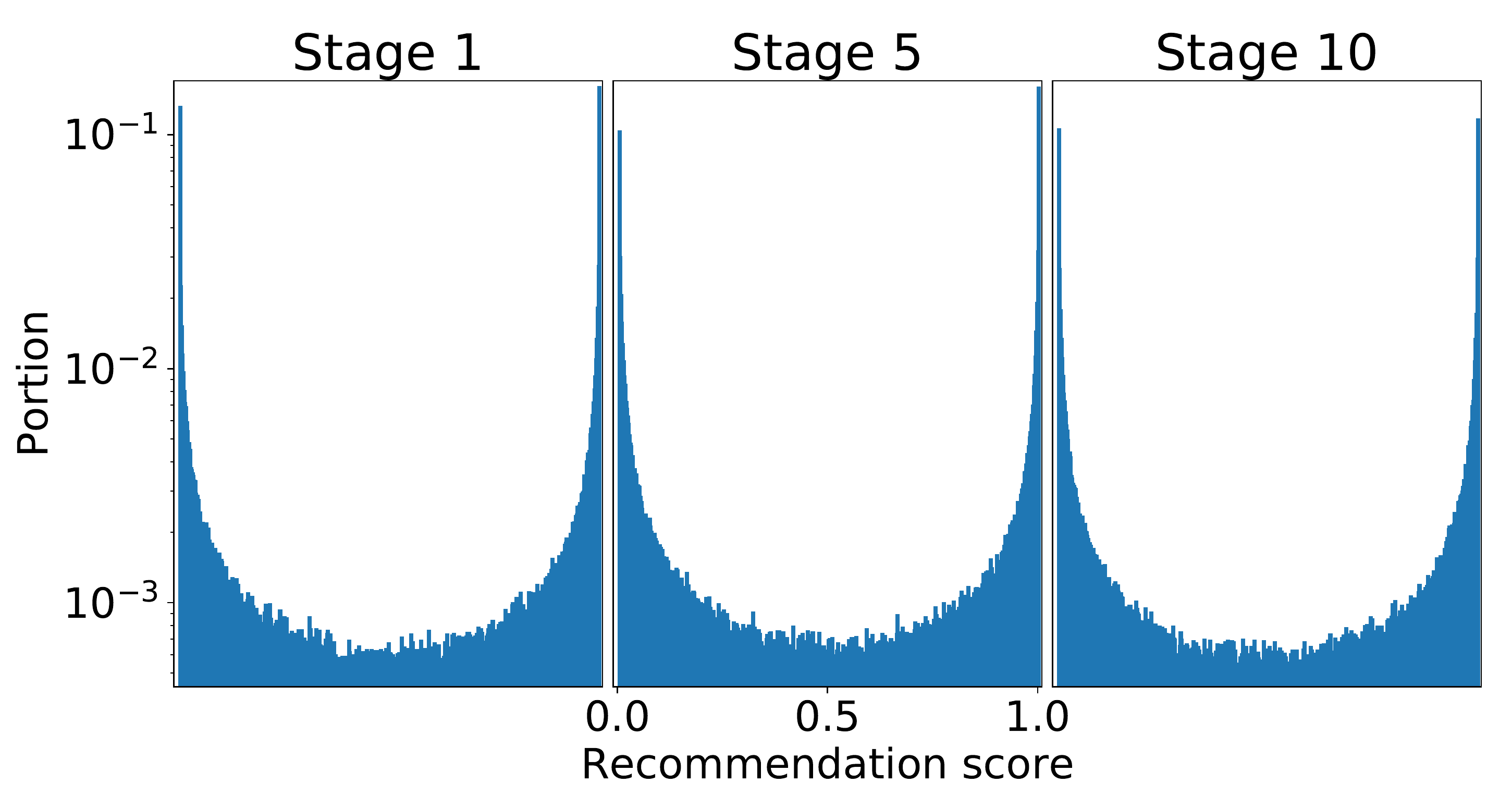}
	\end{center}
	\vspace{-0.2in}
	\caption{Distribution of recommendation scores for boosted regression with NDF. Three stages are visualized and the model is very decisive as the distribution is peaked around 0 and 1.}
	\label{distribution}
\end{figure}
\vspace{-0.05in}
\subsection{Cascaded regression on 3DFAW}

Here we study the decision-making process for a more complex multi-coordinate regression problem on 3DFAW dataset \cite{jeni2016first}. To our best knowledge, this is the first time NDF is boosted and applied on a multi-task regression problem. For an input image $\mathbf{x}_i$, the goal is to predict the position of 66 facial landmarks as a vector $\mathbf{y}_i$. We start with an initialized shape $\hat{\mathbf{y}}_0$ and use a cascade of NDF to update the estimated facial shape stage by stage. The final prediction $\hat{\mathbf{y}} = \hat{\mathbf{y}}_0 + \sum_{t=1}^{K}\Delta \mathbf{y}_{t} $ where $K$ is the total stage number and $\Delta\mathbf{y}_{t}$ is the shape update (model prediction) at stage $ t$. We concatenate 66 local patches cropped around current estimated facial landmarks as input and every leaf node stores a vector as the shape update. We use a cascade length of 10, and in each stage an ensemble of 3 trees is used where each has a depth of 5. The model prediction is shown in Fig.~\ref{result}.

\begin{figure}
	\centering
	\begin{subfigure}[b]{0.55\textwidth}
		\includegraphics[width=0.8\linewidth]{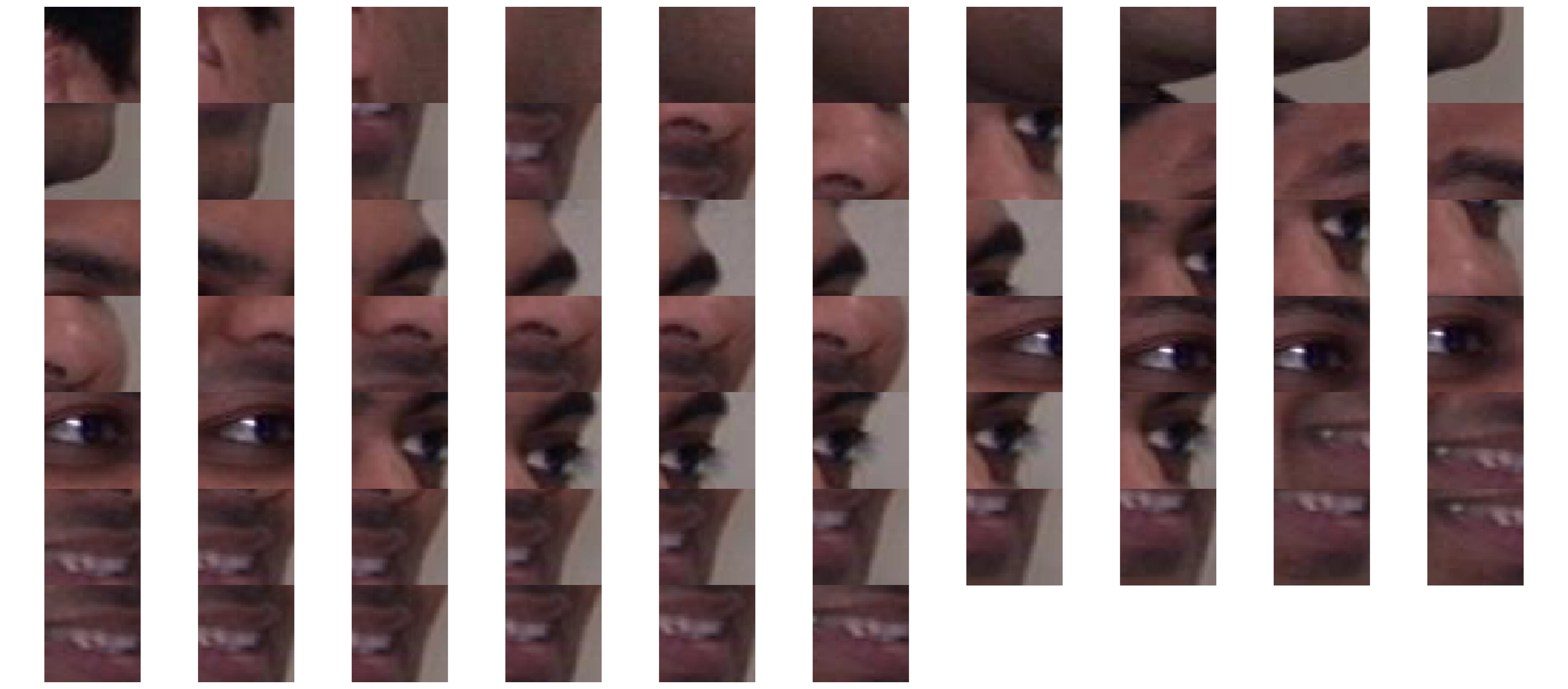}
		\label{fig:Ng1} 
	\end{subfigure}
	
	\begin{subfigure}[b]{0.55\textwidth}
		\includegraphics[width=0.8\linewidth]{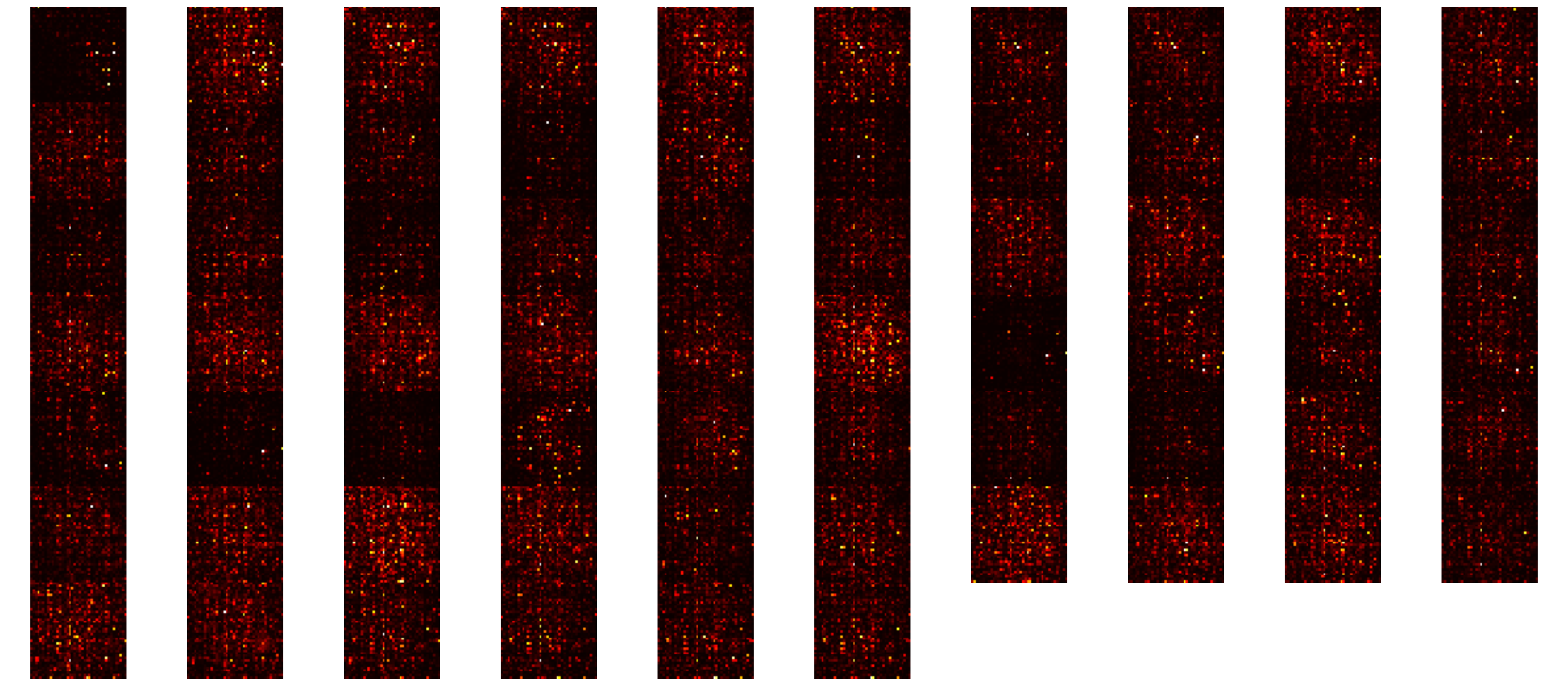}
		\label{fig:Ng2}
	\end{subfigure}
		\caption{Input patches to NDF for regression and their corresponding DSMs.}
		\label{patches}
\end{figure}

The distribution of recommendation scores for this regression problem is shown in Fig.~\ref{distribution}, which is consistent with the results for classification in \cite{7410529}. This means NDF is also decisive for a regression problem and the model can approximate the decision-making process of traditional regression trees. The input patches to the model and their corresponding DSMs for a randomly chosen splitting node are shown in Fig.~\ref{patches}. From these maps we can tell which part of the face influence the decision more during the routing of the input.

\begin{table}
	\begin{center}
		\begin{tabular}{l|c|c}
			\hline
			Dataset & Feature extractor & Accuracy \\
			\hline\hline
			MNIST & Shallow CNN & 99.3\%\\
			CIFAR-10 & VGG16 \cite{DBLP:journals/corr/SimonyanZ14a} & 92.4\%\\
			CIFAR-10 & ResNet50 \cite{he2016deep} & 93.4\% \\
			\hline
		\end{tabular}
	\end{center}
	\vspace{-0.2in}
	\caption{Accuracies for the classification experiments with different feature extractors.}
	\label{accuracy}
\end{table}

\vspace{-0.05in}
\section{Conclusion}
We visualize saliency maps during the decision-making process of NDF for both classification and regression problems to understand with part of the input has larger impact on the model decision. We also apply NDF on a facial landmark regression problem and obtain the distribution of routing probabilities for the first time. The distribution is consistent with the previous classification work and indicates a decisive behavior.

\noindent \textbf{Acknowledgement}. We gratefully acknowledge the support of NVIDIA Corporation with the donation of the Titan Xp GPU used for this research.

{\small
\bibliographystyle{ieee}
\bibliography{reference}

\begin{thebibliography}{10}\itemsep=-1pt

\bibitem{he2016deep}
K.~He, X.~Zhang, S.~Ren, and J.~Sun.
\newblock Deep residual learning for image recognition.
\newblock In {\em Proceedings of the IEEE conference on computer vision and
  pattern recognition}, pages 770--778, 2016.

\bibitem{jeni2016first}
L.~A. Jeni, S.~Tulyakov, L.~Yin, N.~Sebe, and J.~F. Cohn.
\newblock The first 3d face alignment in the wild (3dfaw) challenge.
\newblock In {\em European Conference on Computer Vision}, pages 511--520.
  Springer, 2016.

\bibitem{6909637}
V.~Kazemi and J.~Sullivan.
\newblock One millisecond face alignment with an ensemble of regression trees.
\newblock In {\em 2014 IEEE Conference on Computer Vision and Pattern
  Recognition}, pages 1867--1874, June 2014.

\bibitem{7410529}
P.~Kontschieder, M.~Fiterau, A.~Criminisi, and S.~R. Bulò.
\newblock Deep neural decision forests.
\newblock In {\em 2015 IEEE International Conference on Computer Vision
  (ICCV)}, pages 1467--1475, Dec 2015.

\bibitem{7130626}
S.~Liao, A.~K. Jain, and S.~Z. Li.
\newblock A fast and accurate unconstrained face detector.
\newblock {\em IEEE Transactions on Pattern Analysis and Machine Intelligence},
  38(2):211--223, Feb 2016.

\bibitem{123456}
W.~Shen, Y.~Guo, Y.~Wang, K.~Zhao, B.~Wang, and A.~L. Yuille.
\newblock Deep regression forests for age estimation.
\newblock In {\em The IEEE Conference on Computer Vision and Pattern
  Recognition (CVPR)}, June 2018.

\bibitem{DBLP:journals/corr/SimonyanVZ13}
K.~Simonyan, A.~Vedaldi, and A.~Zisserman.
\newblock Deep inside convolutional networks: Visualising image classification
  models and saliency maps.
\newblock {\em CoRR}, abs/1312.6034, 2013.

\bibitem{DBLP:journals/corr/SimonyanZ14a}
K.~Simonyan and A.~Zisserman.
\newblock Very deep convolutional networks for large-scale image recognition.
\newblock {\em CoRR}, abs/1409.1556, 2014.

\bibitem{8579018}
Q.~{Zhang}, Y.~N. {Wu}, and S.~{Zhu}.
\newblock Interpretable convolutional neural networks.
\newblock In {\em 2018 IEEE/CVF Conference on Computer Vision and Pattern
  Recognition}, pages 8827--8836, June 2018.

\bibitem{DBLP:journals/corr/abs-1802-00121}
Q.~Zhang, Y.~Yang, Y.~N. Wu, and S.~Zhu.
\newblock Interpreting cnns via decision trees.
\newblock {\em CoRR}, abs/1802.00121, 2018.

\end{thebibliography}
}

\end{document}